\documentclass[10pt,journal,compsoc]{IEEEtran}

\usepackage{cite}
\usepackage{amsmath}
\usepackage{amssymb}
\usepackage{graphicx}
\usepackage{booktabs}
\usepackage{array}
\usepackage{multirow}
\usepackage{flushend}
\usepackage{needspace}
\usepackage{xcolor}
\usepackage{url}
\usepackage[hidelinks]{hyperref}

\title{BER-PEF: Unified Human Mobility Predictability Evaluation via Bayes Error Rate Estimation}

\author{En Xu, Jingtao Ding, Zhiwen Yu, and Yong Li%
\IEEEcompsocitemizethanks{%
\IEEEcompsocthanksitem \raggedright E. Xu, J. Ding, and Y. Li are with the Department of Electronic Engineering, Tsinghua University, Beijing, China.
E-mail: xuen@mail.tsinghua.edu.cn, dingjt15@tsinghua.org.cn, and liyong07@tsinghua.edu.cn.
\IEEEcompsocthanksitem Z. Yu is with the School of Computer Science, Northwestern Polytechnical University, Xi'an, China.
E-mail: zhiwenyu@nwpu.edu.cn.}}

\begin{document}
\raggedbottom

\maketitle

\begin{abstract}
Human mobility predictability concerns the best prediction performance attainable from a given target and input information, but its ground truth is not directly observable on real mobility data. We present BER-PEF, a Bayes-error-rate-based framework that converts BER estimation into mobility predictability estimation and provides a unified protocol for comparing estimators without observable ground truth. The framework maps symbolic sequences, numeric trajectories, contextual features, and learned representations into a common feature--label space, then evaluates estimator outputs along controlled perturbation curves against a shared predictability reference interval by measuring deviations below the interval, above the interval, and across the full interval. Experiments on Foursquare NYC and TKY, GeoLife, and T-Drive show that several BER-based estimators achieve lower reference discrepancy than existing predictability methods on symbolic sequences and numeric trajectories, while their estimates track changes in empirical prediction performance under perturbation. Additional analyses show that contextual inputs and multiple structured representations can be evaluated under the same protocol, and that aggregating evidence across multiple perturbation levels provides a more reliable basis for estimator selection than relying on a single unperturbed observation. BER-PEF therefore offers a unified and verifiable path for evaluating predictability estimators on heterogeneous mobility data when ground-truth predictability is unavailable.
\end{abstract}


\section{Introduction}

Mobile spatiotemporal data support urban computing, location-based services, and human behavior modeling. Research on these data considers next-location prediction, point-of-interest (POI) recommendation, and trajectory modeling, and evaluates specific predictors with metrics such as Acc@1, Acc@10, ADE, and FDE~\cite{feng2018deepmove,lian2014geomf,gupta2018socialgan,han2025unimove}. These empirical metrics answer how a particular model performs under a particular train--test protocol. They do not answer how much predictive structure remains after fixing the history, target labels, and representation space, or how far the model is from the attainable limit~\cite{wang2020predictability,xu2026complexsystems,xu2026accuracy}. The latter question belongs to mobility predictability evaluation, which concerns the predictability limit independently of any particular predictor.

Predictability analysis therefore has a practical role beyond reporting another model-specific performance statistic. By providing a task-level reference independent of predictor choice, it helps interpret empirical performance when different algorithms attain different accuracies or alternate as the best performer. Low empirical accuracy relative to a high predictability limit suggests room for improving the predictor or its training procedure, whereas low values of both may indicate limited predictive information under the current observation and representation conditions. Comparing this limit across temporal, semantic, and structural conditions can further reveal which sources of mobility regularity contribute to prediction and inform whether to improve the model, enrich its inputs, or reconsider the task definition~\cite{wang2020predictability,teixeira2021impact,chen2022contrasting,xu2023equivalence,xu2026complexsystems}.

Existing predictability methods provide several routes for estimating this limit. Classical entropy-based methods with Fano-style mappings first estimate entropy from discrete location sequences and then map it to a predictability upper bound; compression-based methods such as Lempel--Ziv exploit repeated sequence structure for entropy estimation~\cite{song2010limits,cover2006elements}. Later studies refine the candidate space or contextual conditions through context transitions, context merging, reachability, and Top-$\ell$ structures~\cite{teixeira2018predictability,chen2016temporal,zhang2022beyond}. Metric-based alternatives, such as permutation entropy, characterize regularity through local ordinal patterns and sequence complexity~\cite{bandt2002permutation}. These methods generally provide a static estimate or a bound under a particular assumption, and remain closely tied to the sequence form, state space, and entropy or complexity definition being used.

\begin{figure*}[!t]
  \centering
  \includegraphics[width=0.96\textwidth]{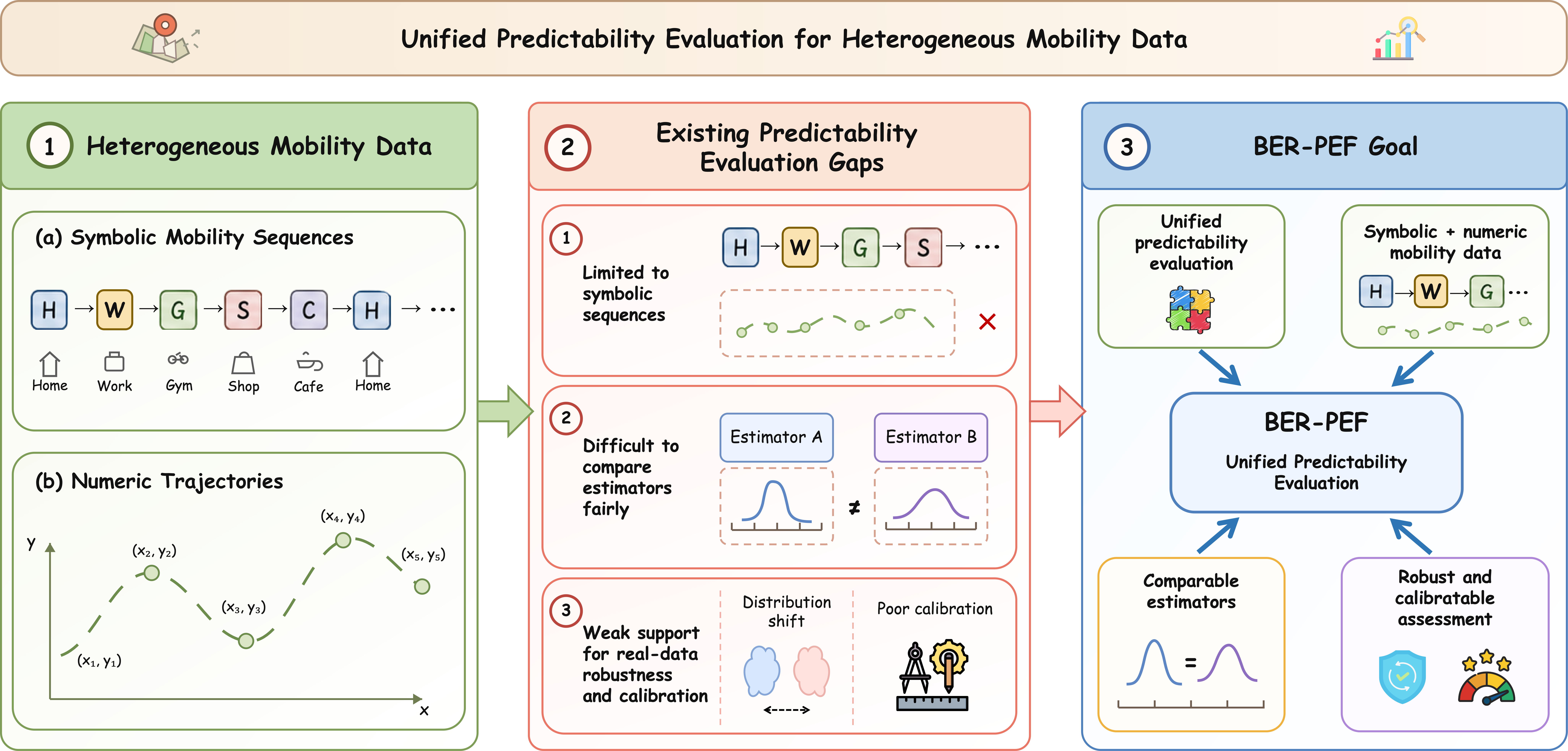}
  \caption{Problem setting for unified human mobility predictability evaluation. Existing methods are difficult to compare on real heterogeneous mobility data because predictability ground truth is unavailable and method assumptions differ across data forms.}
  \label{fig:problem-concept}
\end{figure*}

Two more fundamental issues arise on real mobility data, as illustrated in Fig.~\ref{fig:problem-concept}. First, predictability ground truth is not directly observable. Different estimators may use different state spaces, assumptions, and output conventions~\cite{song2010limits,teixeira2018predictability,bandt2002permutation}, so their static outputs cannot by themselves establish which estimator is reliable. A common protocol is needed in which methods share data, perturbations, representations, reference intervals, and discrepancy metrics. Second, real mobility data include not only discrete check-in sequences but also continuous GPS trajectories, temporal and semantic context, and learned embeddings. Methods tied to a particular symbolic sequence or complexity definition cannot be directly applied to all such inputs. This work therefore asks how to construct and compare predictability estimators on real mobility data without observable ground truth, and how to extend the evaluation to heterogeneous mobility inputs.

We address these issues from the perspective of Bayes error rate (BER). Given a task and a representation space, the predictability limit is the complement of the Bayes-optimal error~\cite{xu2023equivalence}. BER estimation directly targets this optimal error and provides an estimator family covering nearest-neighbor, density, and divergence assumptions~\cite{renggli2021evaluating}. BER-PEF uses this equivalence to turn BER estimators into mobility predictability estimators that produce lower- and upper-side predictability estimates on a common feature--label representation. The BER estimator produces the estimate, whereas the BER-PEF evaluation framework assesses its consistency and reliability.

BER-PEF has two central designs. First, it constructs controlled perturbation curves on the same real data so that estimators share perturbed samples, label spaces, representations, and predictability reference intervals. It then evaluates the entire estimate curve using normalized lower-side, upper-side, and full-interval discrepancies rather than comparing unverifiable static outputs. Second, a unified representation layer maps symbolic sequences, numeric trajectories, and temporal or categorical context to a common feature--label representation, allowing the same BER estimator family to process heterogeneous inputs.

We evaluate BER-PEF on Foursquare NYC, Foursquare TKY, GeoLife, and T-Drive from the perspectives of estimator quality, evaluation reliability, and representation applicability. BER-based predictability estimates achieve low reference discrepancy across symbolic sequences and numeric trajectories and closely track empirical prediction performance under controlled perturbations. Across four datasets, they remain close to and strongly correlated with the best prediction performance observed among a diverse set of algorithms, even as the best-performing algorithm changes across conditions. These results demonstrate the value of predictability as a common task-level reference for interpreting predictor-dependent performance. Multi-point area evaluation provides stronger pooled evidence for estimator selection than an origin-only observation, while contextual inputs and multiple structured representations can be evaluated under the same protocol.

Our contributions are as follows:
\begin{enumerate}
  \item We propose a BER-based approach for estimating human mobility predictability by relating predictability to the complement of Bayes-optimal error. The approach produces predictability estimates for symbolic mobility sequences and continuous trajectories within a common feature--label representation.
  \item We develop a unified comparison framework for predictability estimators on real mobility data. The framework uses controlled perturbation curves, a shared predictability reference interval, and normalized lower-side, upper-side, and full-interval discrepancies to compare estimator outputs and assess reliability using perturbation levels and random realizations not used for estimator selection.
  \item Experiments show that selected BER-based estimates achieve lower reference discrepancy than existing upper-side methods on the evaluated mobility data and track empirical prediction performance under perturbation. These results support BER-based predictability evaluation within the proposed framework.
\end{enumerate}

\section{Related Work}
\label{sec:related}

\subsection{Human Mobility Prediction}

Human mobility prediction studies next-location prediction, trajectory prediction, and POI recommendation from historical locations, time information, user behavior, and spatial context. Early approaches represented mobility as transitions among discrete locations and used Markov chains or probabilistic models such as EPR and TimeGeo to characterize return and exploration patterns~\cite{gambs2012next,jiang2016timegeo}. These models offer clear probabilistic interpretations and low training costs, but their expressive power is constrained by the prescribed transition structure and may not capture high-order spatiotemporal dependencies.

Deep learning subsequently became a major paradigm for mobility prediction, shifting the focus from simple transition probabilities to joint modeling of temporal regularities, periodic preferences, and location interactions. Recurrent and attention-based models capture higher-order dependencies from historical sequences and spatiotemporal context~\cite{liu2016predicting,feng2018deepmove}; long--short-term preference models separately represent stable user preferences and recent behavior~\cite{sun2020where}. Graph, spatiotemporal-attention, and Transformer models further exploit relations among locations, trajectory flows, and non-adjacent visits for next-POI prediction~\cite{luo2021stan,yang2022getnext}. These methods form the main technical line of human mobility prediction, with the common objective of improving empirical predictor accuracy under a specific task and data condition~\cite{luca2021mobilitysurvey}.

As mobility data are increasingly used across cities and regions, recent studies have examined mobility knowledge transfer and unified multi-city prediction to address data scarcity and distribution shifts in new cities~\cite{he2020newcity,jiang2021transfer,han2025unimove}. Some recent work also uses external knowledge to support next-location prediction, but this remains an extension of predictor design~\cite{feng2025agentmove}. In contrast, BER-PEF does not propose a new mobility predictor. It estimates the attainable predictability limit for a given data form, context, and representation, and compares the reliability of predictability estimators under a common protocol.

\subsection{Predictability Evaluation and Bayes Error Estimation}

Predictability research estimates a model-independent prediction limit under given data, task, and representation conditions, providing an attainable reference for algorithm performance~\cite{xu2026complexsystems,xu2026accuracy}. Predictability analysis has also been extended to sequential, top-$N$, and rating prediction in recommender systems~\cite{xu2023quantifying,xu2024limits,xu2025upper}. Classical human mobility studies estimate entropy and combine it with Fano-style bounds to obtain predictability limits for symbolic sequences; compression-based methods such as Lempel--Ziv estimate entropy from repeated sequence structure~\cite{song2010limits,cover2006elements}. Later work revisits this paradigm for high-resolution localization, mobility prediction limits, refined limits, and alternative bounds~\cite{lin2012predictability,lu2013approaching,smith2014refined,ikanovic2017alternative,kulkarni2019examining}, analyzes stationarity, regularity, social sources, and application-collected location data~\cite{teixeira2019deciphering,teixeira2021impact,chen2022contrasting,wang2020predictability}, and incorporates context transitions, context merging, candidate spaces, reachability, context entropy, and Top-$\ell$ structures~\cite{teixeira2018predictability,chen2016temporal,zhang2022beyond}. Metric-based alternatives such as permutation entropy characterize regularity in short, noisy, or numeric time series through sequence complexity~\cite{bandt2002permutation}. These methods provide important foundations, but most target a particular sequence form or report only an upper-side estimate, which complicates unified comparison across continuous trajectories, multidimensional context, and learned embeddings.

Bayes error rate provides a formal bridge between predictability and optimal prediction performance. If predictability denotes the best performance attainable by any classifier in a representation space, BER gives the same limit from the error-rate perspective. BER estimator research covers nearest-neighbor, density/KDE, divergence/GHP, and extrapolation approaches~\cite{cover1967nearest,fukunaga1973nonparametric,fukunaga1987bayes,sekeh2020learning,snapp1995estimating}, and prior work has systematically evaluated these estimators on real classification data~\cite{renggli2021evaluating}. The equivalence between time-series predictability and Bayes error has also been established formally, with controlled theoretical models illustrating how BER estimation can address limitations of entropy-based predictability estimates~\cite{xu2023equivalence}. These results do not, however, provide a complete workflow for real mobility spatiotemporal data that combine symbolic sequences, continuous trajectories, multidimensional context, and learned representations while lacking observable predictability ground truth. BER-PEF extends this connection to real mobility evaluation through a unified representation layer, controlled perturbation curves, and a common protocol for quantifying estimator reliability.

\section{The BER-PEF Framework}
\label{sec:method}

\begin{figure*}[!t]
  \centering
  \includegraphics[width=0.95\textwidth]{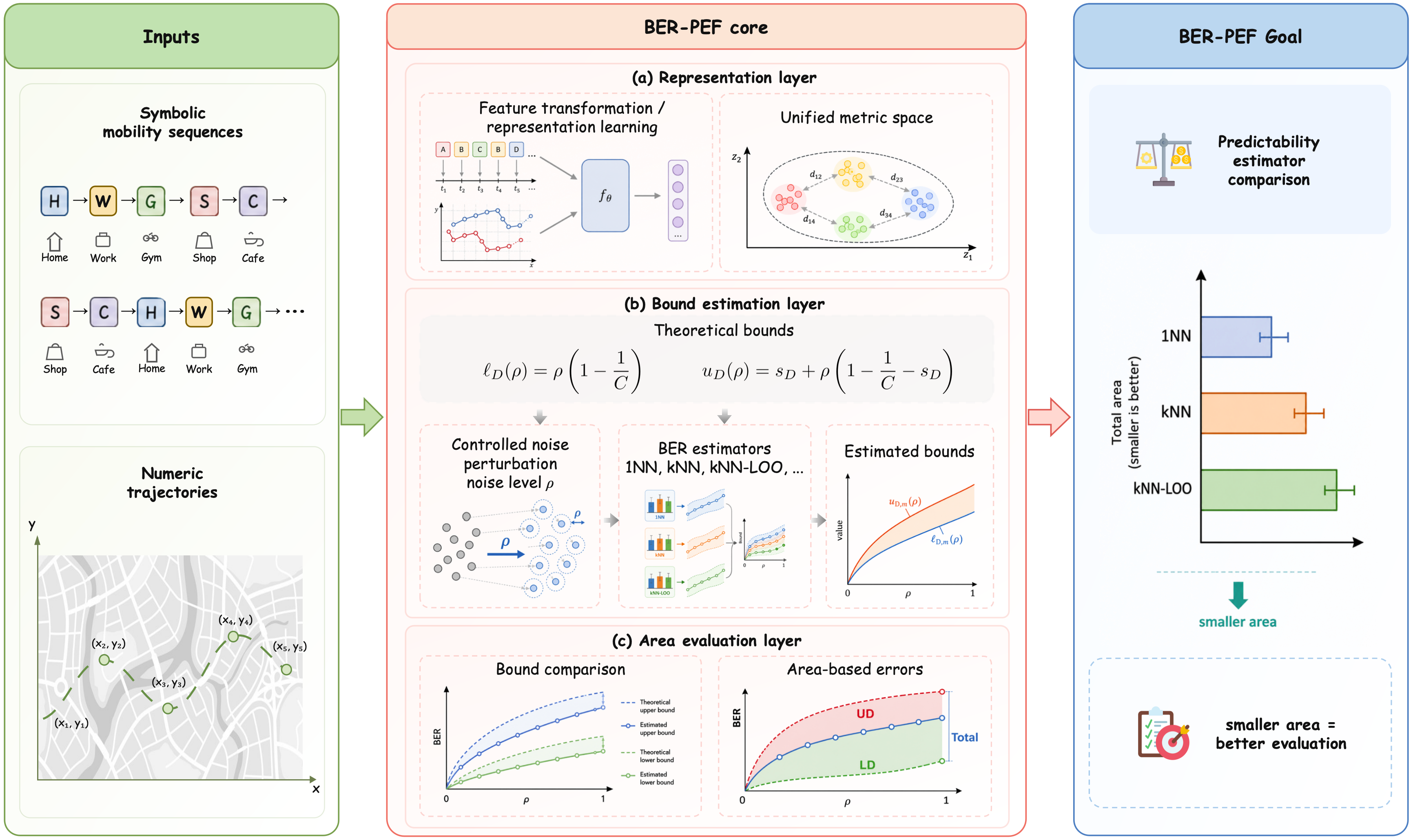}
  \caption{Overview of BER-PEF. A common representation layer supports symbolic and numeric mobility inputs; a bound-estimation layer applies BER estimators under controlled perturbations; and an area-evaluation layer compares estimated curves with a shared predictability reference interval.}
  \label{fig:core-framework}
\end{figure*}

\subsection{Problem Formulation}

Human mobility predictability describes how accurately a target mobility state can, in principle, be predicted from a given history and representation. It is a property jointly determined by the data-generating process, observable information, and representation space, rather than the empirical accuracy of a particular predictor under one train--test protocol. We formalize predictability as the optimal attainable prediction performance in a given representation space.

Raw mobility data are typically user-ordered temporal sequences rather than independent samples. Let the mobility sequence of user $u$ be
\begin{equation}
S_u=\{(o_{u,1},c_{u,1}), (o_{u,2},c_{u,2}), \ldots, (o_{u,T_u},c_{u,T_u})\},
\end{equation}
where $o_{u,t}$ is the $t$-th mobility observation, such as a discrete location, continuous coordinate, or trajectory state, and $c_{u,t}$ denotes optional context such as time or location category. To convert next-step predictability into the feature--label form required by BER estimators, BER-PEF constructs a history window of length $L$:
\begin{equation}
\begin{aligned}
h_{u,t}^{(L)}
&=\bigl((o_{u,\tau},c_{u,\tau})\bigr)_{\tau=t-L}^{t-1},\\
x_{u,t}
&=\psi(h_{u,t}^{(L)}),\qquad y_{u,t}=g(o_{u,t}).
\end{aligned}
\end{equation}
Here, $\psi$ organizes the history window as model input, and $g$ maps the next observation to a class label. For symbolic sequences, $g(o_{u,t})$ typically denotes a next-location or location-cluster ID; for continuous trajectories, it denotes a discretized next-state class. Context enters through $h_{u,t}^{(L)}$ rather than acting as a separate prediction target, allowing us to quantify the effect of additional information on predictability.

Let the resulting mobility prediction dataset be defined over input space $\mathcal{X}$ and label space $\mathcal{Y}=\{1,\ldots,C\}$, where $C$ is the number of target classes. Re-indexing all valid $(u,t)$ samples gives
\begin{equation}
D=\{(x_j,y_j)\}_{j=1}^{n}, \qquad (x_j,y_j)\sim P_D(X,Y),
\end{equation}
where $x_j\in\mathcal{X}$ is a history-window input and $y_j\in\mathcal{Y}$ is its next-step target. A representation function
\begin{equation}
\phi:\mathcal{X}\rightarrow\mathbb{R}^{d}, \qquad z_j=\phi(x_j)
\end{equation}
maps heterogeneous mobility inputs to $d$-dimensional numeric representations. Thus, $\phi$ is the common interface through which BER-PEF accepts symbolic sequences, continuous trajectories, and contextual features. A BER estimator receives
\begin{equation}
D_{\mathrm{BER}}=\{(z_j,y_j)\}_{j=1}^{n}.
\end{equation}
Once the window construction $\psi$, label mapping $g$, and representation $\phi$ are fixed, predictability on a mobility sequence becomes an optimal classification-error estimation problem in feature--label space:
\begin{equation}
\text{predictability of } S_u
\quad \Longleftrightarrow \quad
\text{Bayes error of } D_{\mathrm{BER}}.
\end{equation}

Given $\phi$, a mobility predictor is a function $f:\mathbb{R}^{d}\rightarrow\mathcal{Y}$. Its empirical error on $D$ is
\begin{equation}
\widehat{R}_D(f;\phi)=\frac{1}{n}\sum_{j=1}^{n}\mathbf{1}\{f(\phi(x_j))\neq y_j\}.
\end{equation}
The commonly reported Acc@1 equals $1-\widehat{R}_D(f;\phi)$ and therefore measures the empirical performance of one predictor under the current protocol, not the predictability limit of the data and representation space.

We instead consider the optimal attainable error in $\phi(\mathcal{X})$. Let the population risk of predictor $f$ be
\begin{equation}
R_D(f;\phi)=\Pr_{(X,Y)\sim P_D}\left[f(\phi(X))\neq Y\right].
\end{equation}
The Bayes-optimal error in this representation space is
\begin{equation}
R_D^*(\phi)=\inf_f R_D(f;\phi),
\end{equation}
and the corresponding predictability is
\begin{equation}
\Pi_D(\phi)=1-R_D^*(\phi).
\end{equation}
Estimating $\Pi_D(\phi)$ is therefore equivalent to estimating $R_D^*(\phi)$. We use the error-rate formulation because $R_D^*(\phi)$ directly corresponds to the Bayes error rate and connects naturally to existing BER estimators and theoretical bounds~\cite{xu2023equivalence,renggli2021evaluating}.

BER-PEF does not train a new next-location predictor. It evaluates whether a predictability estimator reliably estimates $\Pi_D(\phi)$ or its equivalent error-rate form. For estimator $m$, a static estimate can be denoted by $\widehat{\Pi}_{D,m}(\phi)$. Under the controlled perturbation protocol introduced below, we extend it to lower- and upper-side predictability curves $\widehat{\Pi}_{D,m}^{\mathrm{L}}(\rho)$ and $\widehat{\Pi}_{D,m}^{\mathrm{U}}(\rho)$ and compare estimator reliability over perturbation strength $\rho$.

\subsection{Unified Evaluation Architecture}

BER-PEF maps heterogeneous mobility data into a common representation space, uses BER estimators to construct lower- and upper-side predictability curves under controlled perturbations, and compares estimator reliability through normalized area discrepancies.

As shown in Fig.~\ref{fig:core-framework}, BER-PEF has three layers. The \textbf{representation layer} converts symbolic sequences, numeric trajectories, and temporal or categorical context into numeric features. The \textbf{bound-estimation layer} applies BER estimators to perturbed samples and produces lower- and upper-side predictability estimates. The \textbf{area-evaluation layer} compares these curves with a predictability reference interval and summarizes their discrepancies over perturbation strength.

\subsubsection{Unified Representation Layer}

\begin{figure*}[!t]
  \centering
  \includegraphics[width=0.8\textwidth]{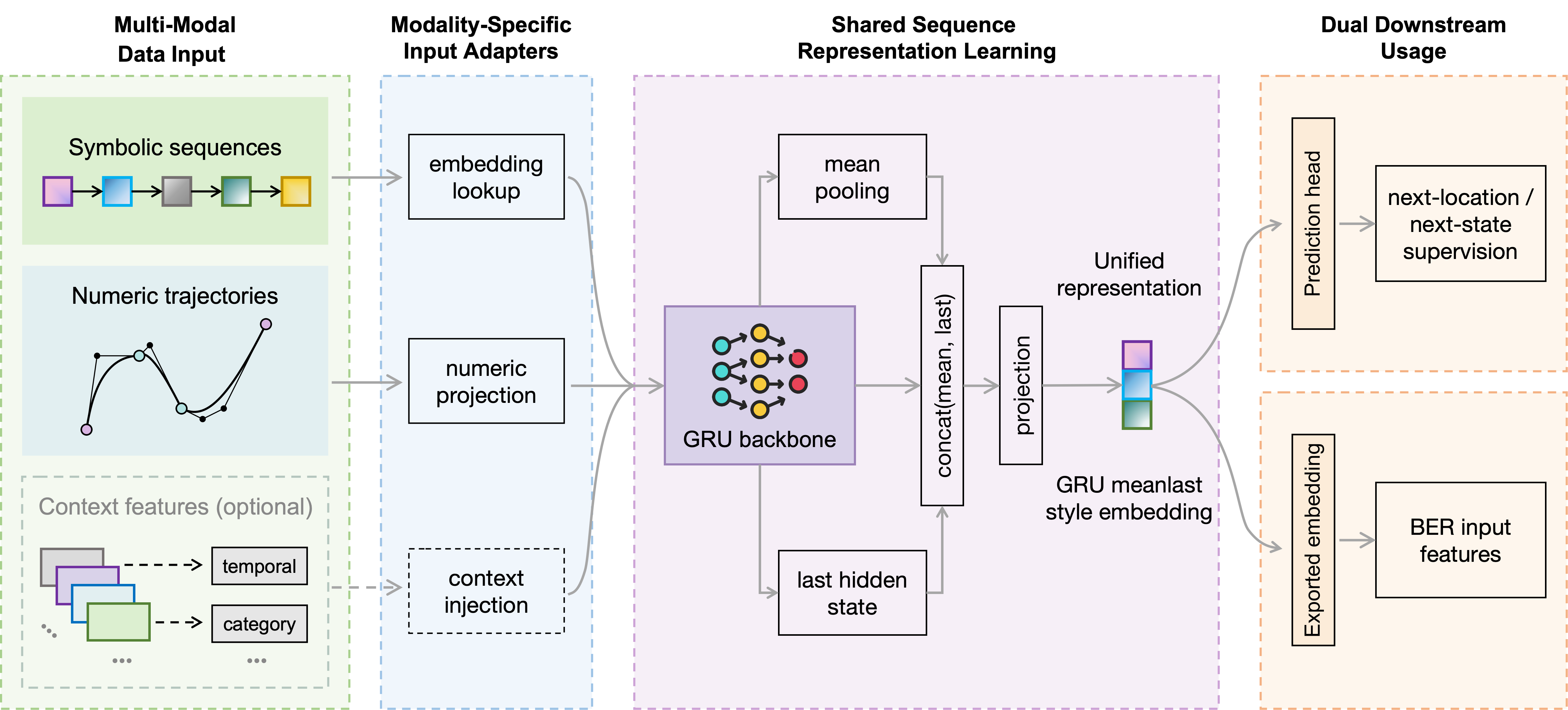}
  \caption{Unified representation module in BER-PEF. Symbolic, numeric, and contextual inputs are adapted to a shared sequence encoder; temporal aggregation and projection produce the sample representation used with its target label for BER estimation.}
  \label{fig:representation-module}
\end{figure*}

Figure~\ref{fig:representation-module} details the representation layer. Input adapters map symbolic sequences, numeric trajectories, and optional context into a common sequence representation. A shared sequence encoder, mean--last temporal aggregation, and linear projection then produce a sample representation $z_j$. The prediction head supplies next-step supervision during representation learning, whereas BER estimation operates on the resulting feature--label samples.

Formally, for history-window input $x_j$, the adapters define
\begin{equation}
r_{j,1:L}=\alpha(x_j), \qquad r_{j,\tau}\in\mathbb{R}^{p}.
\end{equation}
For a symbolic main input, a lookup table maps each discrete location or state ID to a dense vector:
\begin{equation}
e_{j,\tau}^{\mathrm{sym}}=E_{\mathrm{loc}}[o_{j,\tau}^{\mathrm{sym}}].
\end{equation}
For a numeric main input, a linear projection maps normalized coordinates, displacements, or local motion states $\tilde{o}_{j,\tau}^{\mathrm{num}}$ into the same vector space:
\begin{equation}
e_{j,\tau}^{\mathrm{num}}=W_{\mathrm{num}}\tilde{o}_{j,\tau}^{\mathrm{num}}+b_{\mathrm{num}}.
\end{equation}
These two branches define $e_{j,\tau}^{\mathrm{main}}\in\{e_{j,\tau}^{\mathrm{sym}},e_{j,\tau}^{\mathrm{num}}\}$. If temporal, categorical, or other context $c_{j,\tau}$ is available, a context embedding gives
\begin{equation}
e_{j,\tau}^{\mathrm{ctx}}=E_{\mathrm{ctx}}[c_{j,\tau}],
\end{equation}
and $e_{j,\tau}^{\mathrm{ctx}}=\mathbf{0}$ when no context is used. The input to the shared encoder is
\begin{equation}
r_{j,\tau}=W_r[e_{j,\tau}^{\mathrm{main}};e_{j,\tau}^{\mathrm{ctx}}]+b_r .
\end{equation}
These operations correspond to the embedding lookup, numeric projection, and context injection in Fig.~\ref{fig:representation-module}. The adapted sequence is processed by
\begin{equation}
H_j=E_\theta(r_{j,1:L})=(h_{j,1},\ldots,h_{j,L}),
\end{equation}
where $E_\theta$ is the shared sequence encoder. Our primary representation, \textsc{UniMob}, uses a recurrent sequence backbone.

The temporal aggregation in \textsc{UniMob} combines mean pooling and the last hidden state:
\begin{align}
H_j &= \mathrm{GRU}_\theta(r_{j,1:L}), \\
a_j &= \left[h_{j,L};\frac{1}{L}\sum_{\tau=1}^{L}h_{j,\tau}\right], \\
z_j &= P(a_j)=W_p a_j+b_p .
\end{align}
The full representation function is
\begin{equation}
\phi_\theta(x_j)=P(A(E_\theta(\alpha(x_j))))=z_j.
\end{equation}
Thus, $z_j$ is the output of the representation function $\phi_\theta$ for sample $x_j$ and determines the sample's coordinates in the representation space $\phi_\theta(\mathcal{X})$. Subsequent BER estimation operates on the feature--label pair $(z_j,y_j)$ rather than on the encoder's intermediate state $H_j$ or the prediction head output $\hat{p}_j$.

To preserve discriminative structure relevant to mobility prediction, the representation is trained using next-location or next-state supervision. Let $q_\omega$ denote the prediction head. Then
\begin{equation}
\hat{p}_j=\mathrm{softmax}(q_\omega(z_j)),\quad
\mathcal{L}_{\mathrm{sup}}
=-\frac{1}{n}\sum_{j=1}^{n}\log \hat{p}_{j,y_j}.
\end{equation}
This objective learns features related to next-step label $y_j$, but BER-PEF does not use $q_\omega$, $\hat{p}_j$, or their accuracy as the final predictability estimate. The learned representation is paired with $y_j$ to form
\begin{equation}
D_{\mathrm{BER}}=\{(z_j,y_j)\}_{j=1}^{n}.
\end{equation}
\textsc{UniMob} is the primary implementation of the representation layer, not a mandatory component of BER-PEF. Different choices of $\phi_\theta$ define different representation spaces; the subsequent task is to determine whether an estimator reliably estimates $R_D^*(\phi_\theta)$ or its bounds in the chosen space.

\subsubsection{Predictability-Curve Evaluation Protocol}

The fundamental challenge on real mobility data is that the true predictability limit is unobservable. Estimator accuracy therefore cannot be judged against a known label as in ordinary supervised learning. A single estimate on unperturbed data also cannot reveal whether differences arise from task difficulty or estimator behavior. BER-PEF instead constructs a controlled difficulty curve on real data. As perturbation strength $\rho$ increases, exploitable regularity should systematically decrease. A reliable predictability estimator should follow this change and remain close to a reference interval constrained jointly by theoretical randomness and achieved model performance.

For each dataset $D$ and perturbation strength $\rho$, BER-PEF perturbs the same original samples and requires every estimator to share the perturbed instances, label space, and representation input. This turns each dataset into a predictability evaluation curve from low to high perturbation. Estimator comparison then tests whether an estimated curve respects the predictability constraints induced by this controlled change in difficulty. Section~\ref{sec:experimental-setup} specifies the perturbation levels, repetitions, and data configurations.

Let $C$ be the number of classes in $D$ and $a_D^{\mathrm{SOTA}}$ the SOTA accuracy on the original task. BER-PEF defines two predictability-side reference bounds:
\begin{align}
\Pi_D^{\mathrm{L}}(\rho)
&=(1-\rho)a_D^{\mathrm{SOTA}}+\rho\frac{1}{C}, \\
\Pi_D^{\mathrm{U}}(\rho)
&=1-\rho\left(1-\frac{1}{C}\right).
\end{align}
These are reference quantities rather than estimator outputs. The conservative lower reference $\Pi_D^{\mathrm{L}}(\rho)$ starts from the demonstrated performance $a_D^{\mathrm{SOTA}}$ and degrades toward random-guess accuracy $1/C$ as perturbations remove learnable structure. The upper reference $\Pi_D^{\mathrm{U}}(\rho)$ captures the fact that even an optimal predictor cannot recover the randomized fraction, and therefore declines according to the random-error rate $1-1/C$. Together they define the interval $[\Pi_D^{\mathrm{L}}(\rho),\Pi_D^{\mathrm{U}}(\rho)]$ within which a reasonable predictability estimate should lie. They are complements of the corresponding error-rate references; we use the predictability-side formulation throughout the evaluation.

For estimator $m$, let its predictability-side outputs at strength $\rho$ be
\begin{equation}
\widehat{\Pi}_{D,m}^{\mathrm{L}}(\rho), \qquad
\widehat{\Pi}_{D,m}^{\mathrm{U}}(\rho).
\end{equation}
Native error-rate bounds are complemented to obtain these quantities. If $\widehat{\Pi}_{D,m}^{\mathrm{L}}(\rho)$ falls below $\Pi_D^{\mathrm{L}}(\rho)$, the estimator regards the task as harder than demonstrated model performance supports, producing a lower-side underestimate. If $\widehat{\Pi}_{D,m}^{\mathrm{U}}(\rho)$ exceeds $\Pi_D^{\mathrm{U}}(\rho)$, it assigns more regularity than is permitted by the randomized fraction, producing an upper-side overestimate.

Let $d(x,[l,u])=\max\{l-x,0,x-u\}$ be the distance from $x$ to interval $[l,u]$, and let $I_D(\rho)=[\Pi_D^{\mathrm{L}}(\rho),\Pi_D^{\mathrm{U}}(\rho)]$ denote the reference interval. BER-PEF aggregates deviations over $[0,0.9]$ using normalized areas:
\begin{align}
\mathrm{LD}_{D,m}
&=\frac{1}{0.9}\int_0^{0.9}d\!\left(\widehat{\Pi}_{D,m}^{\mathrm{L}}(\rho),I_D(\rho)\right)d\rho, \\
\mathrm{UD}_{D,m}
&=\frac{1}{0.9}\int_0^{0.9}d\!\left(\widehat{\Pi}_{D,m}^{\mathrm{U}}(\rho),I_D(\rho)\right)d\rho, \\
\mathrm{Total}_{D,m}
&=\mathrm{LD}_{D,m}+\mathrm{UD}_{D,m}.
\end{align}
The discrete implementation first aggregates repeated runs at each $\rho$ and then applies trapezoidal integration over the actual $\rho$ coordinates rather than equally averaging nonuniform observation levels. LD and UD evaluate the lower- and upper-side estimates, respectively, while Total measures full-interval discrepancy; lower values indicate closer agreement with the reference interval. If two estimators have identical discrepancy at $\rho=0$ but one has a strictly lower discrepancy over an interval of positive measure, an origin-only protocol cannot distinguish them whereas an area protocol can. This property shows that area evaluation retains interval information; its practical reliability advantage is tested separately using held-out data.

\subsection{BER Estimator Family and Extensibility}

BER-PEF treats predictability estimators as interchangeable modules rather than binding the framework to one BER estimator. Given feature--label data $D_{\mathrm{BER}}^{(\rho)}$ and class count $C$, every $m\in\mathcal{M}$ follows the interface
\begin{equation}
m\colon(D_{\mathrm{BER}}^{(\rho)},C)\mapsto
(\widehat{\Pi}_{D,m}^{\mathrm{L}}(\rho),
\widehat{\Pi}_{D,m}^{\mathrm{U}}(\rho)),\quad m\in\mathcal{M}.
\end{equation}
All estimators are thus compared under the same embedding representation, perturbed samples, label space, reference interval, and area metrics. BER-PEF contributes a common mobility predictability evaluation protocol for existing BER estimators rather than prescribing one fixed estimator~\cite{renggli2021evaluating}.

Our experiments use three core estimators: \textbf{1NN($k=3$)}, \textbf{kNN}, and \textbf{kNN-LOO}. The first forms a local BER estimate from nearest-neighbor structure, kNN characterizes local classification uncertainty through neighborhood label distributions, and kNN-LOO reduces self-matching bias through leave-one-out evaluation. These estimators are related to classical nearest-neighbor bounds and kNN estimates of Bayes risk and Bayes error~\cite{cover1967nearest,devijver1985multiclass,devroye1981asymptotic,fukunaga1975knn}. The extended family includes \textbf{GHP}, \textbf{KDE}, \textbf{KDE-kNN-LOO}, \textbf{kNN-Ext}, and \textbf{LR} to cover additional assumptions. GHP uses a multiclass Bayes error bound based on generalized Henze--Penrose divergence~\cite{sekeh2020learning}; KDE and KDE-kNN-LOO represent density/Parzen and kNN posterior-estimation approaches~\cite{fukunaga1973nonparametric,fukunaga1987bayes}; and kNN-Ext extrapolates Bayes risk from nearest-neighbor sample statistics~\cite{snapp1995estimating}. The extended family demonstrates estimator compatibility and is not assumed to outperform the core estimators.

The interface also permits future BER estimators to join $\mathcal{M}$ if they accept $D_{\mathrm{BER}}^{(\rho)}$ and output predictability-side curves. They can then be compared under the same perturbations, reference interval, and LD/UD/Total metrics. SOTA-reference robustness and empirical performance calibration are external evaluation evidence for the estimator family, not part of the estimator interface itself.

\section{Experiments}
\label{sec:experiments}

\subsection{Experimental Setup}
\label{sec:experimental-setup}

We evaluate BER-PEF across symbolic mobility sequences and continuous trajectories using a common perturbation-based reference convention. The experiments compare estimator agreement, empirical calibration, task-level prediction benchmarks, evaluation reliability, and representation robustness; the protocol specific to each research question is introduced with its corresponding results.

\subsubsection{Datasets and Tasks}

We use two forms of real mobility data. Symbolic experiments use Foursquare NYC and Foursquare TKY~\cite{yang2015modeling}, where each sample contains a history of discrete location IDs and the label is the next-location class. Numeric experiments use GeoLife~\cite{zheng2011geolife} and T-Drive~\cite{yuan2010tdrive,yuan2011driving}; trajectory points are represented in local coordinates and discretized into next-state prediction tasks. These datasets cover common check-in sequence and continuous trajectory settings and test whether one protocol can compare predictability across data forms.

For controlled comparisons across datasets, representations, and estimators, we construct a fixed-size evaluation subset for each dataset and apply representation learning, perturbation generation, and curve evaluation to that subset. The subset is the controlled comparison unit of the protocol, not a survey sample intended to estimate a city-wide user distribution. User filtering and per-user sample caps control computational scale and sample density across sources. Windowing places symbolic sequences and numeric trajectories in a common supervised interface; our objective is to evaluate predictability estimators in a fixed representation space rather than optimize long-history mobility predictors. Table~\ref{tab:dataset-stats} reports the resulting sample counts and task settings. The quantity $a_D^{\mathrm{SOTA}}$ is the original-task SOTA accuracy used to construct the reference interval.

\begin{table*}[!t]
\centering
\caption{Dataset statistics and task settings.}
\label{tab:dataset-stats}
\setlength{\tabcolsep}{8pt}
\begin{tabular}{lllrrr}
\toprule
Dataset & Data form & Task & Train/Test samples & $C$ & $a_D^{\mathrm{SOTA}}$ \\
\midrule
Foursquare NYC & Symbolic & Next location & 12,780 / 1,801 & 497 & 0.6130 \\
Foursquare TKY & Symbolic & Next location & 13,967 / 2,076 & 466 & 0.6267 \\
GeoLife & Numeric & Next state & 38,656 / 4,804 & 200 & 0.7161 \\
T-Drive & Numeric & Next state & 38,315 / 4,930 & 200 & 0.4254 \\
\bottomrule
\end{tabular}
\end{table*}

\subsubsection{Compared Methods}

The first comparison group provides task-specific predictability references. On symbolic mobility sequences, we evaluate three entropy-based methods. \textbf{LZ-Fano} estimates an LZ entropy rate and maps it to predictability through a Fano-style relation~\cite{song2010limits,cover2006elements}. \textbf{Reachability-Fano} uses the same entropy estimate and mapping but replaces the global class count with the empirical reachability size $N_r=\max_x|\mathcal{N}(x)|$, where $\mathcal{N}(x)$ is the set of distinct observed successors of state $x$~\cite{smith2014refined,teixeira2019deciphering}. \textbf{Permutation Predictability} estimates sequence regularity from ordinal-pattern complexity~\cite{bandt2002permutation}. For continuous trajectories, we use \textbf{PermEnt-2D}, a two-dimensional permutation-entropy method. Context analysis additionally includes \textbf{Predictability(CTX)} and \textbf{Predictability(CTX-Merge)}, corresponding to context-transition and context-merge variants~\cite{teixeira2021impact,zhang2022beyond}. Their applicable inputs and assumptions differ across data forms.

The second group contains BER-PEF estimators. The core family comprises 1NN($k=3$), kNN, and kNN-LOO. The extended analysis includes GHP, KDE, KDE-kNN-LOO, kNN-Ext, and LR, covering nearest-neighbor, density/KDE, divergence/GHP, and extrapolation approaches~\cite{cover1967nearest,devijver1985multiclass,fukunaga1973nonparametric,fukunaga1987bayes,sekeh2020learning,snapp1995estimating,renggli2021evaluating}. Every BER-PEF estimator uses the same perturbed samples, representations, label space, and reference interval. For existing methods that produce only an upper-side predictability estimate, we report UD. For BER-PEF estimators that produce both sides of the interval, we report LD, UD, and Total.

\subsection{Cross-Form Estimator Comparison under a Unified Perturbation Protocol}
\label{sec:main-results}

\begin{figure*}[!t]
  \centering
  \includegraphics[width=0.84\textwidth,trim=0 13.5bp 0 0,clip]{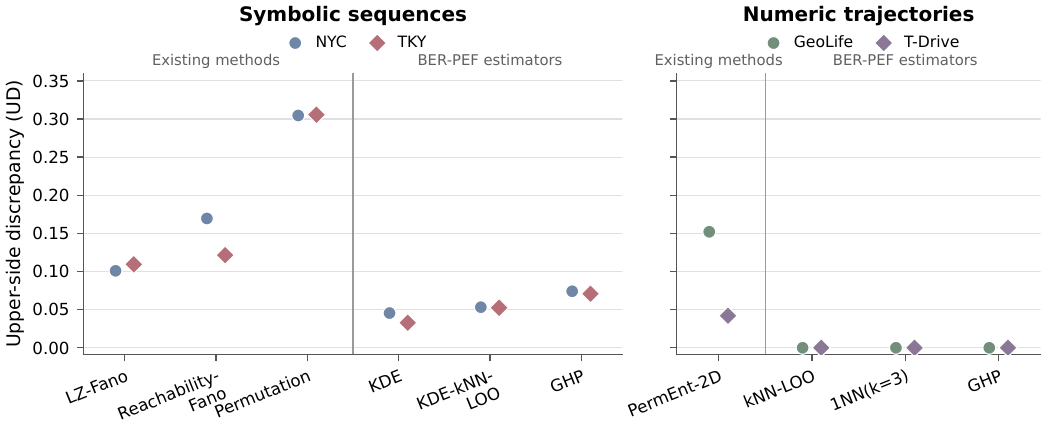}
  \setlength{\abovecaptionskip}{3pt}
  \caption{Upper-side discrepancy of existing predictability methods and the top-three BER-PEF estimators selected by mean Total across the two datasets within each data form. The left and right panels compare symbolic sequences and numeric trajectories on a shared UD scale; lower values indicate closer agreement with the upper-side reference.}
  \label{fig:main-ud-comparison}
\end{figure*}

\begin{figure*}[!t]
  \centering
  \includegraphics[width=0.80\textwidth,trim=0 7.5bp 0 0,clip]{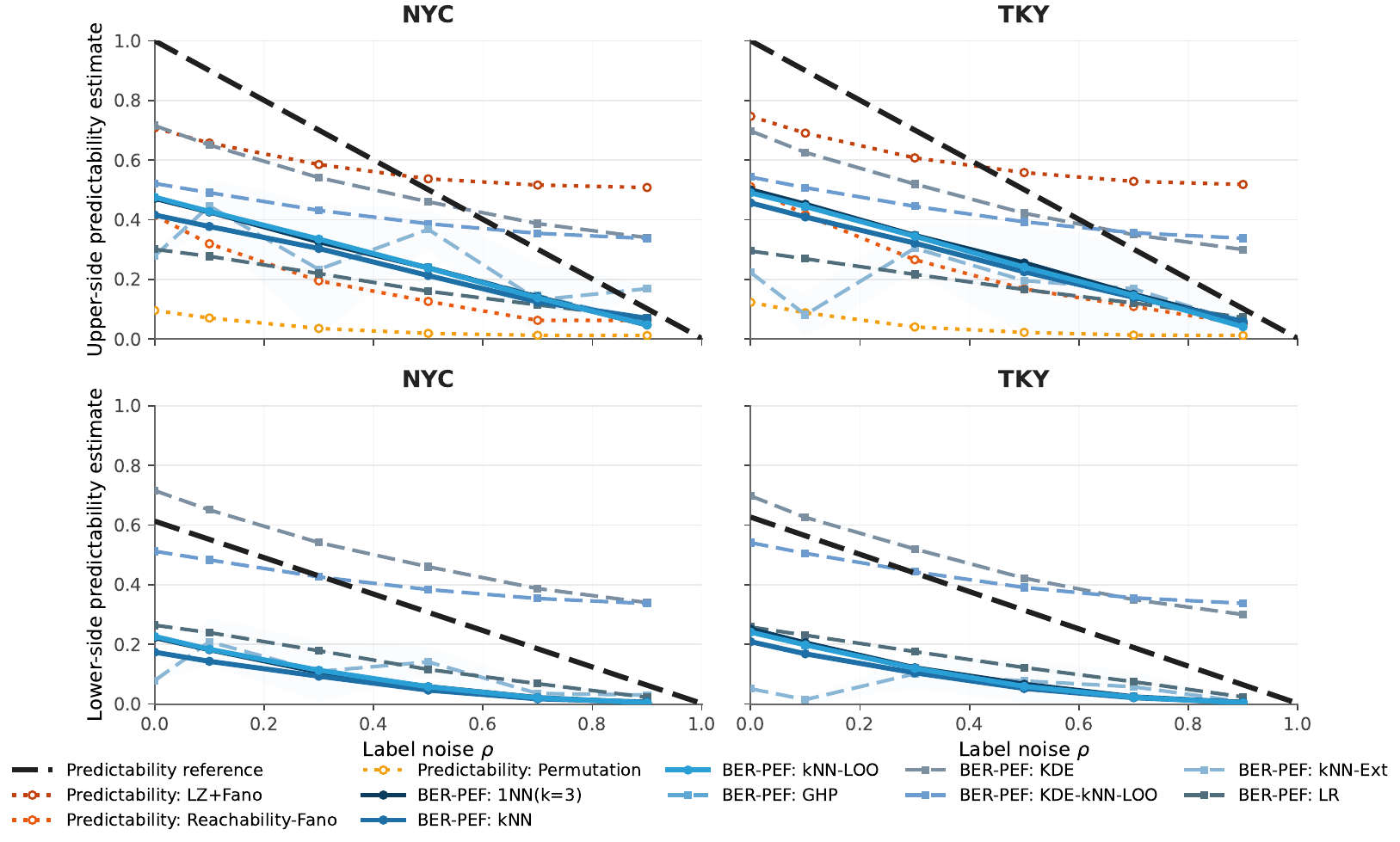}
  \setlength{\abovecaptionskip}{3pt}
  \caption{Predictability estimation curves on symbolic mobility sequences.}
  \label{fig:symbolic-main}
\end{figure*}

This section examines the reference agreement of BER-based estimators across heterogeneous mobility data. Existing predictability methods that output only an upper-side estimate are evaluated with upper-side discrepancy (UD). BER-based estimators that produce both sides of the reference interval are additionally evaluated with lower-side discrepancy (LD) and Total discrepancy ($\mathrm{Total}=\mathrm{LD}+\mathrm{UD}$). UD measures agreement with the upper reference, whereas Total measures aggregate deviation over the full perturbation interval. Figure~\ref{fig:main-ud-comparison} summarizes UD on four datasets, and the complete perturbation curves are shown in Figs.~\ref{fig:symbolic-main} and~\ref{fig:numeric-main}.

Figure~\ref{fig:main-ud-comparison} summarizes the upper-side comparison on four datasets. To avoid selecting methods from an individual dataset or from the plotted UD itself, we first rank the formal BER estimator family by mean Total over the two datasets within each data form and then fix the top three. The selected estimators are KDE, KDE-kNN-LOO, and GHP for symbolic sequences, and kNN-LOO, 1NN($k=3$), and GHP for numeric trajectories. Selection is therefore based on full-interval quality, whereas the figure reports UD to enable comparison with existing methods that provide only an upper-side estimate.

\subsubsection{Symbolic Mobility Sequences}

For symbolic mobility sequences, LZ-Fano is the strongest existing predictability method on NYC and TKY, with UD values of $0.101$ and $0.109$, respectively. Reachability-Fano obtains $0.170$ and $0.121$, while Permutation Predictability obtains $0.305$ and $0.306$. The KDE, KDE-kNN-LOO, and GHP estimators selected by mean Total obtain UD values of $0.046/0.033$, $0.053/0.052$, and $0.074/0.071$ on NYC/TKY, respectively, all below the strongest existing method on the corresponding city.

Some BER estimators provide closer upper-reference agreement on discrete location sequences, although quality varies across the family. KDE and KDE-kNN-LOO obtain Total values of $0.091/0.066$ and $0.109/0.105$, respectively. GHP has low UD but Total values of $0.325/0.322$, indicating that its main discrepancy arises on the lower side. BER-PEF therefore compares upper-side estimates while also exposing full-interval deviations that are invisible under UD alone.

\subsubsection{Numeric Trajectories}

For numeric trajectories, BER-PEF also achieves close reference agreement. PermEnt-2D obtains UD values of $0.152$ on GeoLife and $0.042$ on T-Drive. The kNN-LOO, 1NN($k=3$), and GHP estimators selected by mean Total obtain UD values close to zero on both datasets. Their Total values on GeoLife are $0.073$, $0.075$, and $0.075$, respectively, and their values on T-Drive are $0.035$, $0.035$, and $0.036$.

\begin{figure*}[!t]
  \centering
  \includegraphics[width=0.74\textwidth,trim=0 7.5bp 0 0,clip]{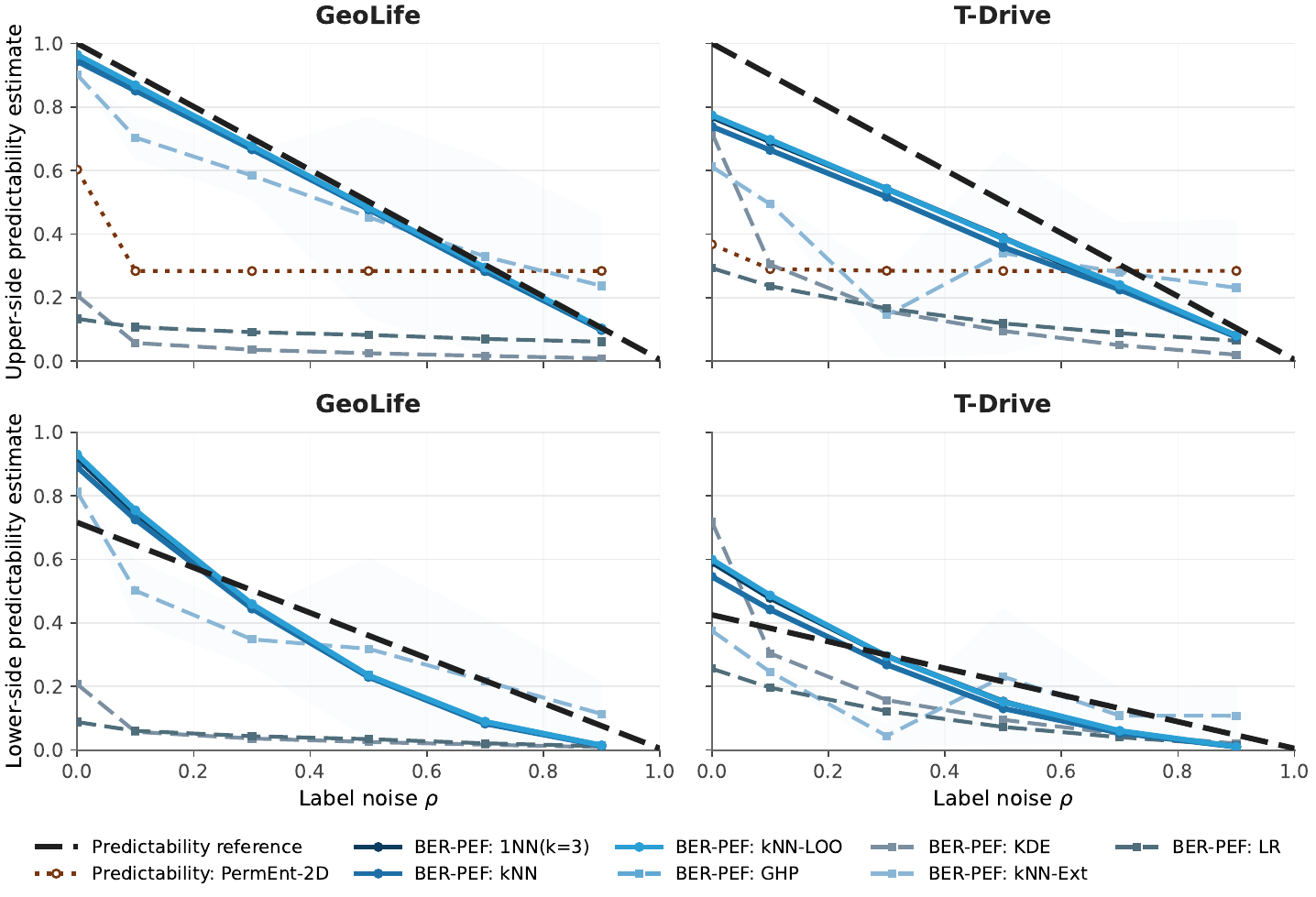}
  \setlength{\abovecaptionskip}{3pt}
  \caption{Predictability estimation curves on numeric trajectories.}
  \label{fig:numeric-main}
\end{figure*}

The three estimators therefore achieve both near-zero UD and low full-interval discrepancy on continuous trajectories. The curves in Fig.~\ref{fig:numeric-main} further show that BER-based estimates follow the reference interval as trajectory perturbation increases. These results indicate that BER-PEF can compare estimator quality for discrete sequences and continuous trajectory representations under one reference-interval protocol.

\subsection{Calibration to Empirical Prediction Performance}
\label{sec:empirical-calibration}

This experiment asks whether predictability estimates are calibrated to observed prediction performance in both trend and numerical scale. On Foursquare NYC and TKY, two independent runs at $\rho\in\{0,0.1,0.3,0.5,0.7,0.9\}$ yield 12 aligned tasks per dataset. For each dataset--run--$\rho$ key, Linear-Ngram, \textsc{SeqMLP}, and \textsc{UniMob} use the same perturbed labels, and their highest Acc@1 defines an empirical best-of-three proxy. BER lower error estimates are mapped to upper-side predictability as $1-\widehat{R}_{\mathrm{lower}}$. We measure trend agreement using Pearson and Spearman correlations, goodness of linear fit using OLS $R^2$, and scale calibration using MAE to the identity line $y=x$; aggregate values weight the two datasets equally.

Predictability(LZ) follows changes in the proxy but remains systematically above the identity line, with a dataset-balanced Pearson correlation of $0.9575$ and identity MAE of $0.4372$. Among the evaluated BER estimators, BER-LR is the best calibrated, attaining Pearson correlation $0.9986$, $R^2=0.9972$, and identity MAE $0.0111$; BER-1NN($k=3$) and BER-kNN-LOO also achieve Pearson correlations above $0.997$. This contrast distinguishes trend agreement from scale calibration: strong correlation alone does not ensure that an estimate matches the magnitude of observed prediction performance.

These results reinforce the preceding comparison based on the reference interval: the best-performing BER estimators not only track changes in prediction performance but also match its observed scale substantially more closely than Predictability(LZ). This demonstrates the empirical calibration value of BER-based predictability estimation for mobility prediction tasks.

\begin{figure*}[!t]
  \centering
  \includegraphics[width=0.88\textwidth,trim=0 6.5bp 0 0,clip]{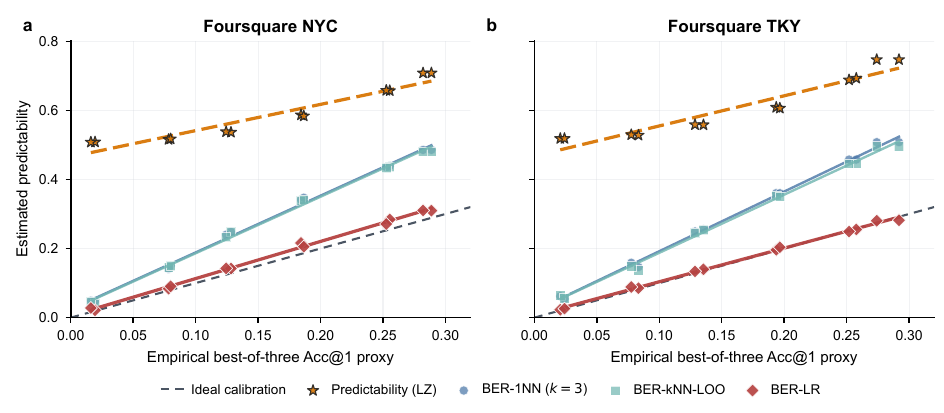}
  \setlength{\abovecaptionskip}{3pt}
  \caption{Calibration of predictability estimates to the empirical best-of-three Acc@1 proxy on Foursquare NYC and TKY. The proxy is the highest Acc@1 among Linear-Ngram, \textsc{SeqMLP}, and \textsc{UniMob} for each aligned task; markers denote task instances, solid lines are OLS fits, and the dashed diagonal denotes exact calibration.}
  \label{fig:empirical-calibration}
\end{figure*}

\subsection{BER Predictability against an Empirical Algorithm Benchmark}
\label{sec:empirical-benchmark-stability}

This experiment examines whether a BER-based predictability estimate can closely and consistently track empirical prediction performance across data forms. We evaluate Foursquare NYC, Foursquare TKY, GeoLife, and T-Drive at six perturbation levels $\mathcal{R}=\{0,0.1,0.3,0.5,0.7,0.9\}$, with ten independent train--test perturbation realizations per level. BER-kNN and the seven prediction algorithms shared by all four datasets---1NN, 5NN, Logistic, Linear SVM, Random Forest, Extra Trees, and MLP---use the same samples, labels, split, and noise seed for every task key. Let $\mathcal{D}$ denote the four datasets, $d\in\mathcal{D}$ a dataset, $r\in\{1,\ldots,10\}$ a realization, and $\rho\in\mathcal{R}$ a perturbation level. For dataset $d$, let $\mathcal{A}_d$ be its complete applicable algorithm pool, $a\in\mathcal{A}_d$ an algorithm, and $\mathrm{Acc@1}_{d,r,\rho}(a)$ its accuracy on the corresponding task. We define the empirical best-of-family benchmark as
\begin{equation}
B_{d,r,\rho}
=
\max_{a\in\mathcal{A}_d}\mathrm{Acc@1}_{d,r,\rho}(a),
\end{equation}
the highest observed Acc@1 for dataset $d$, realization $r$, and perturbation level $\rho$.

\begin{figure*}[!t]
  \centering
  \includegraphics[width=0.95\textwidth,trim=0 11.5bp 0 0,clip]{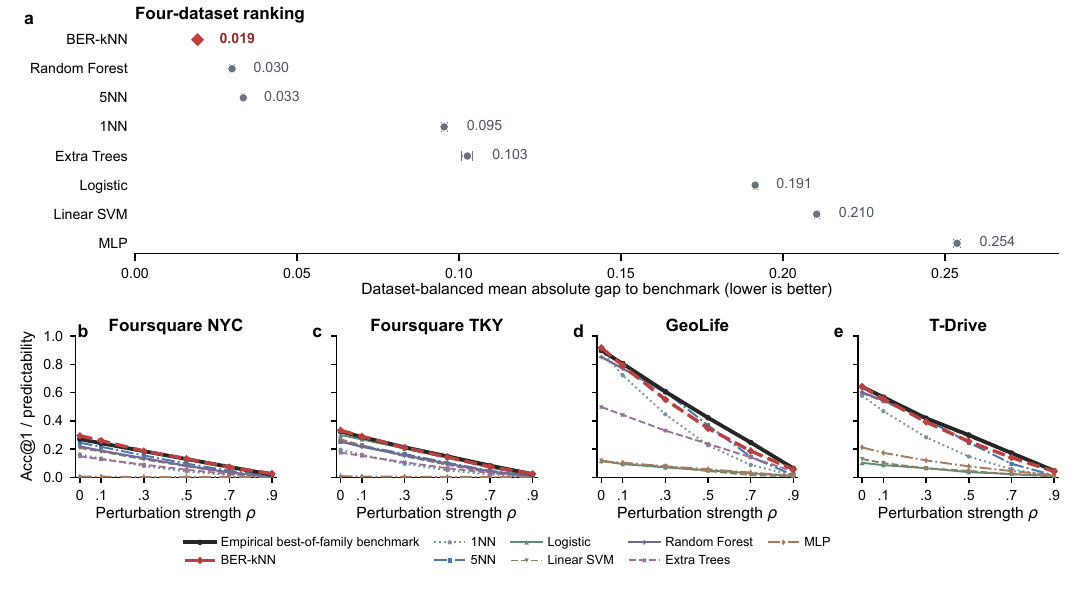}
  \setlength{\abovecaptionskip}{3pt}
  \caption{Comparison with the empirical best-of-family benchmark: (a) dataset-balanced mean absolute gap for BER-kNN and the seven prediction algorithms shared by all datasets; (b--e) mean perturbation curves on Foursquare NYC, Foursquare TKY, GeoLife, and T-Drive. Error bars in (a) and shading for the benchmark and BER-kNN in (b--e) denote 95\% CIs over ten realizations.}
  \label{fig:empirical-benchmark-stability}
\end{figure*}

The ranking compares BER-kNN with the seven shared prediction algorithms. For a ranked method $m$, define its performance-scale value $P_{d,r,\rho}(m)$ as $\mathrm{Acc@1}_{d,r,\rho}(m)$ for a prediction algorithm and $1-\widehat{R}^{\mathrm{kNN}}_{d,r,\rho}$ for BER-kNN. The dataset-level mean absolute gap $G_{d,m}$ and its four-dataset macro average $S_m$ are
\begin{align}
G_{d,m}
&=\frac{1}{10|\mathcal{R}|}\sum_{r=1}^{10}\sum_{\rho\in\mathcal{R}}
\left|P_{d,r,\rho}(m)-B_{d,r,\rho}\right|, \\
S_m&=\frac{1}{|\mathcal{D}|}\sum_{d\in\mathcal{D}} G_{d,m},
\end{align}
respectively. Figure~\ref{fig:empirical-benchmark-stability}(a) and Table~\ref{tab:common-method-ranking} report the ranking by $S_m$, while Fig.~\ref{fig:empirical-benchmark-stability}(b)--(e) shows the perturbation curves. Uncertainty is quantified by two-sided Student-$t$ 95\% confidence intervals over the ten independent realizations.

\begin{table*}[!t]
\centering
\caption{Mean absolute gap to the empirical best-of-family benchmark across four mobility datasets (lower is better).}
\label{tab:common-method-ranking}
\small
\setlength{\tabcolsep}{8pt}
\begin{tabular}{lrrrrr}
\toprule
Method & NYC & TKY & GeoLife & T-Drive & Macro \\
\midrule
BER-kNN & 0.011425 & 0.007166 & 0.038392 & 0.020209 & \textbf{0.019298} \\
Random Forest & 0.042541 & 0.050979 & 0.014776 & 0.011487 & 0.029946 \\
5NN & 0.025412 & 0.046203 & 0.033993 & 0.028093 & 0.033425 \\
1NN & 0.080835 & 0.092718 & 0.107983 & 0.100166 & 0.095425 \\
Extra Trees & 0.078817 & 0.090575 & 0.224830 & 0.016176 & 0.102600 \\
Logistic & 0.002952 & 0.014387 & 0.445108 & 0.303181 & 0.191407 \\
Linear SVM & 0.047983 & 0.051228 & 0.446645 & 0.295740 & 0.210399 \\
MLP & 0.150120 & 0.175233 & 0.439745 & 0.249801 & 0.253725 \\
\bottomrule
\end{tabular}
\end{table*}

BER-kNN ranks first with a four-dataset mean gap of $0.019298$, followed by Random Forest ($0.029946$) and 5NN ($0.033425$). Its gaps on NYC, TKY, GeoLife, and T-Drive are $0.011425$, $0.007166$, $0.038392$, and $0.020209$, respectively. BER-kNN therefore provides the closest task-level approximation to the empirical best-of-family benchmark among the eight common methods.

Across the six perturbation-level means, Pearson correlation between BER-kNN and the benchmark ranges from $0.995$ to $0.999$ over the four datasets, and Spearman correlation is $1.0$ throughout. BER-kNN thus consistently follows the decline in empirical prediction performance as perturbation increases, demonstrating stable agreement in both magnitude and trend.

More importantly, this experiment demonstrates the direct diagnostic value of predictability analysis. The stable agreement between BER-kNN and the empirical algorithm envelope across data forms and perturbation conditions provides a consistent reference for the task-level prediction limit. A substantial gap between an algorithm and the BER predictability estimate indicates remaining room for algorithmic or representational improvement, whereas convergence of the empirical best-performing algorithm toward the estimate suggests that further gains are increasingly constrained by the predictable information available in the data. Predictability estimation therefore not only characterizes task difficulty but also helps distinguish algorithm mismatch from data-limited prediction performance.

\subsection{Selection Reliability of Multi-Point Area Evaluation}

Observing only the unperturbed point may overlook estimator deviations that emerge as perturbation increases. Across symbolic NYC/TKY, numeric GeoLife/T-Drive, and context-enhanced NYC/TKY, we fix 1NN($k=3$), kNN, and kNN-LOO and partition a 19-level grid into candidate levels $\mathcal{R}_P=\{0.1i\}_{i=0}^{9}$ and held-out levels $\mathcal{R}_H=\{0.05+0.1i\}_{i=0}^{8}$. Origin-only selects the estimator with the lowest Total at $\rho=0$, whereas the prespecified protocol $\{0,0.1,0.3,0.4,0.5,0.6,0.8,0.9\}$ selects by normalized Total area. Selection uses ten repetitions on $\mathcal{R}_P$; validation uses 20 disjoint repetitions on $\mathcal{R}_H$.

\begin{figure*}[!t]
  \centering
  \includegraphics[width=0.82\textwidth,trim=0 7.5bp 0 0,clip]{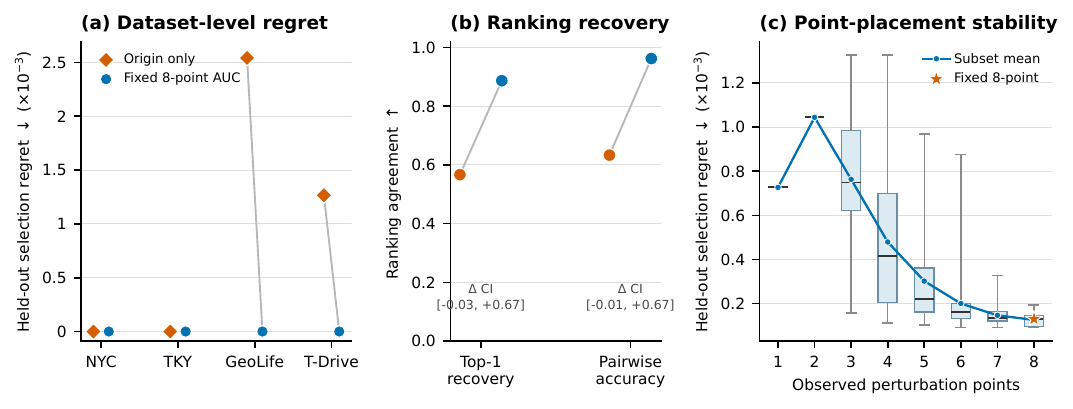}
  \setlength{\abovecaptionskip}{3pt}
  \caption{Estimator-selection reliability of multi-point area evaluation. (a) Held-out regret for Origin-only and the fixed eight-point protocol (lower is better); (b) pooled top-1 recovery and pairwise ordering accuracy, with 95\% hierarchical bootstrap intervals for the protocol difference; (c) subset regret distributions for one to eight observation points.}
  \label{fig:protocol-ablation}
\end{figure*}

The primary metric is held-out selection regret, defined as the Total difference between the selected estimator and the held-out oracle; regret equal to zero means that the protocol selected the held-out oracle. Figures~\ref{fig:protocol-ablation}(a)--(b) show that, in the pooled evaluation over six symbolic, numeric, and context conditions, the eight-point protocol reduces regret from $0.000727$ to $0.000130$, increases oracle top-1 recovery from $0.567$ to $0.887$, and increases pairwise ordering accuracy from $0.633$ to $0.962$. The difference between the eight-point and Origin-only regret is $-0.000596$, with a 95\% hierarchical bootstrap interval of $[-0.001476,0.000085]$; this difference is directionally consistent with the improvements in pooled regret, top-1 recovery, and pairwise ordering accuracy.

Figure~\ref{fig:protocol-ablation}(c) enumerates all endpoint-anchored observation subsets with two to eight points. As the number of points increases from four to six and eight, mean subset regret decreases from $0.000479$ to $0.000201$ and $0.000126$, while worst-case regret decreases from $0.001326$ to $0.000875$ and $0.000193$. These results show that multi-point area evaluation reduces estimator-selection sensitivity to both the single origin and the specific placement of observation levels, providing interval-level evidence for selection.

\subsection{Predictive Information and Estimator Quality with Contextual Features}
\label{sec:context-analysis}

\begin{figure*}[!t]
  \centering
  \includegraphics[width=0.80\textwidth,trim=0 4.5bp 0 0,clip]{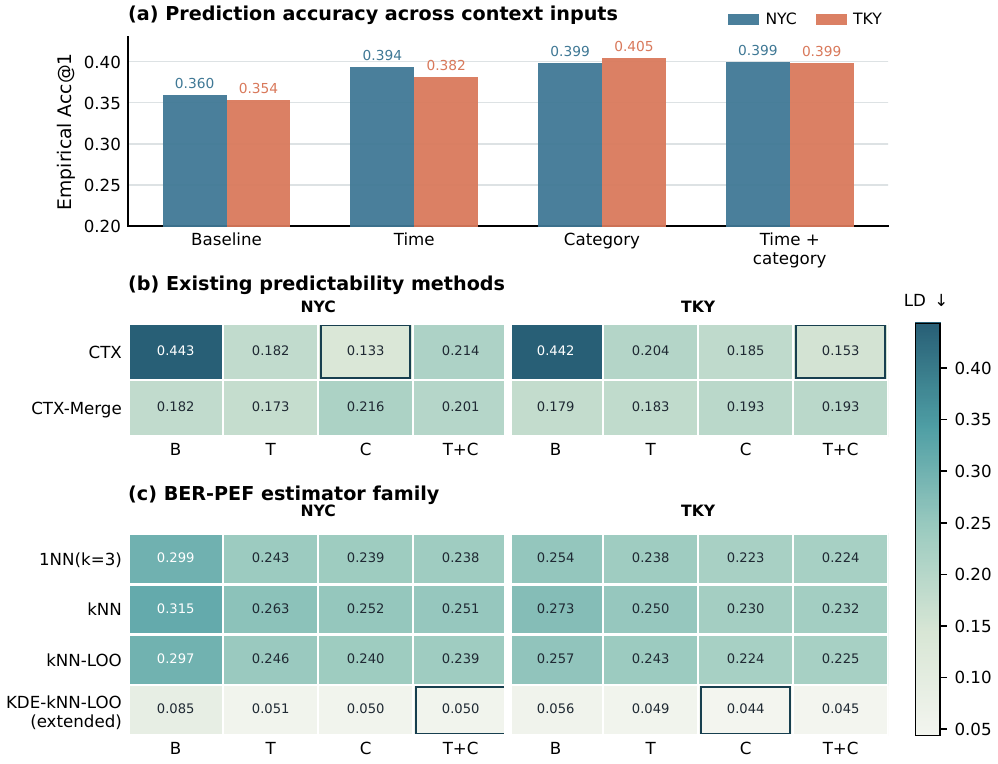}
  \setlength{\abovecaptionskip}{3pt}
  \caption{Context information and estimator quality on NYC and TKY. (a) Empirical Acc@1 under baseline, time, category, and time-plus-category inputs (higher is better); (b) LD of context-aware predictability methods; (c) LD of the BER estimator family. The two LD panels share one color scale, and lower values are better.}
  \label{fig:context-analysis}
\end{figure*}

We evaluate contextual inputs from two perspectives. Empirical Acc@1 measures whether temporal and location-category features provide additional predictive information, whereas LD measures the agreement between a predictability estimator and the lower-side reference for a given contextual representation. On NYC and TKY, we compare baseline, time, category, and time-plus-category inputs while fixing the history length, sample size, sequence-encoder and prediction-head architectures, training protocol, perturbation protocol, and reference definition. Figure~\ref{fig:context-analysis}(a) shows that all three contextual inputs improve empirical prediction performance: Acc@1 increases from $0.360$ to $0.394$--$0.399$ on NYC and from $0.354$ to $0.382$--$0.405$ on TKY. Category reaches $0.405$ on TKY, while category and time-plus-category both reach approximately $0.399$ on NYC. Temporal and location-category context therefore contain information useful for next-location prediction.

Figures~\ref{fig:context-analysis}(b)--(c) compare CTX, CTX-Merge, the three core BER estimators, and the extended estimator KDE-kNN-LOO under the same LD metric. All methods can be evaluated on baseline, time, category, and time-plus-category inputs. The best LD of the two context-aware predictability methods ranges from $0.133$ to $0.201$ over the eight dataset--context conditions, whereas KDE-kNN-LOO obtains $0.044$--$0.085$ and is lower in every condition. Each core estimator also has lower LD after adding context than under its corresponding baseline. BER-PEF therefore supports context-enhanced inputs under one reference convention and can identify estimators with closer lower-side agreement. This comparison illustrates how predictability analysis can be used to assess the value of additional information. The improvement in empirical prediction performance shows that temporal and location-category context provides useful information for next-location prediction, while the corresponding BER evaluation places this improvement in a task-level framework rather than attributing it only to a particular predictor.

\subsection{Prediction Effectiveness and BER Evaluation Robustness of Unified Representations}

We examine the unified representation layer from two perspectives: whether a representation retains information needed for next-location prediction, and whether BER-PEF provides a consistent assessment of estimator quality in that representation. Prediction effectiveness compares Linear-Ngram, \textsc{SeqMLP}, and \textsc{UniMob} on NYC/TKY using Acc@1, Acc@10, and cosine-retrieval Hit@10. BER quality covers symbolic, numeric, and time-plus-category context representations. The core estimator family is fixed to 1NN($k=3$), kNN, and kNN-LOO, and the lowest discrepancy within this family is reported for each available dataset--representation setting. This best-of-core value represents the BER quality attainable under the representation; Total always comes from one estimator and satisfies $\mathrm{Total}=\mathrm{LD}+\mathrm{UD}$.

\begin{figure*}[!t]
  \centering
  \includegraphics[width=0.80\textwidth,trim=0 9.5bp 0 0,clip]{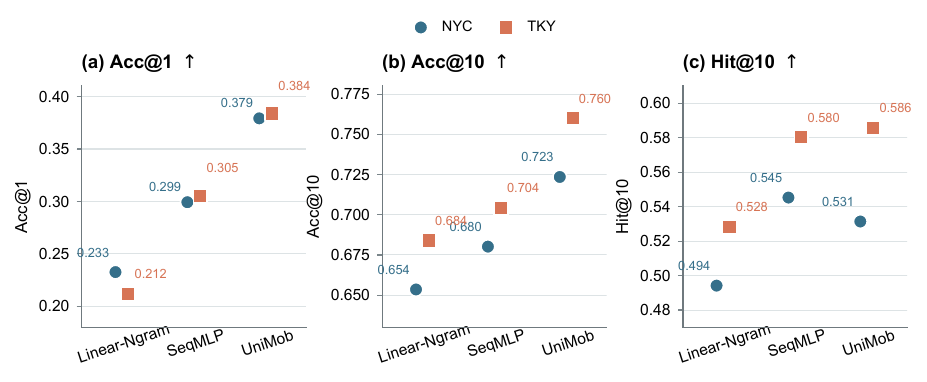}
  \setlength{\abovecaptionskip}{3pt}
  \caption{Representation quality on NYC/TKY symbolic mobility sequences. (a)--(b) Acc@1 and Acc@10 under the same supervised protocol; (c) Hit@10 for cosine retrieval with test queries and a training-set gallery. Higher values are better, and all source values are annotated.}
  \label{fig:representation-prediction-quality}
\end{figure*}

Figure~\ref{fig:representation-prediction-quality} shows that \textsc{UniMob} obtains Acc@1 values of $0.379/0.384$ on NYC/TKY, above the $0.299/0.305$ of \textsc{SeqMLP} and $0.233/0.212$ of Linear-Ngram. Its Acc@10 values of $0.723/0.760$ are also highest in both cities. Hit@10 supplements these measures by assessing local neighborhood structure: \textsc{UniMob} reaches the highest value on TKY ($0.586$), while its $0.531$ on NYC is close to the $0.545$ of \textsc{SeqMLP}. These results show that \textsc{UniMob} retains effective next-location prediction information and competitive local neighborhood structure.

\begin{figure*}[!t]
  \centering
  \includegraphics[width=0.86\textwidth,trim=0 7.5bp 0 0,clip]{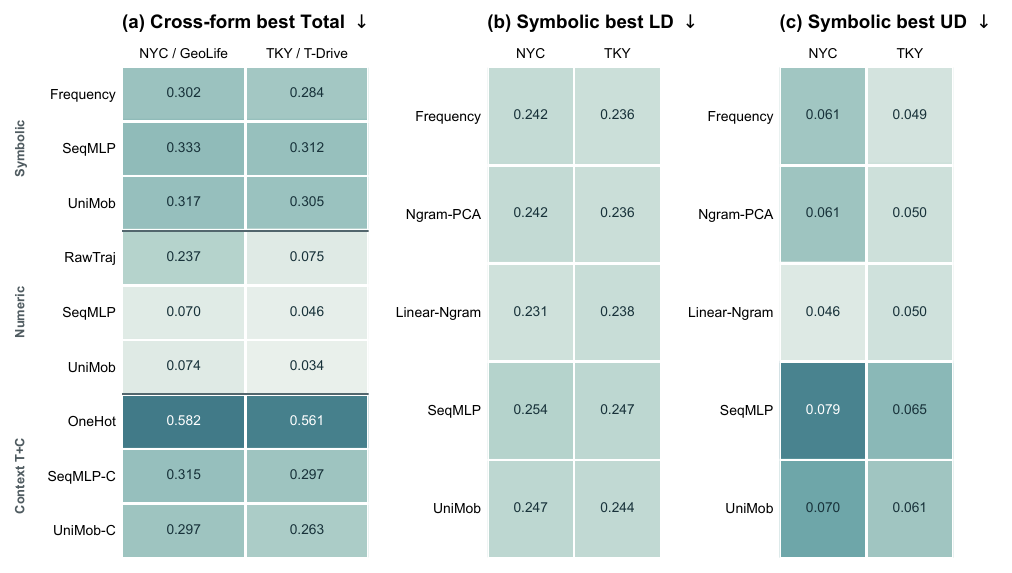}
  \setlength{\abovecaptionskip}{3pt}
  \caption{Lowest discrepancy attainable by the predefined core BER family under different representations (lower is better). (a) Cross-form best Total for symbolic, numeric, and fixed time-plus-category context inputs; (b)--(c) best LD and best UD for five symbolic representations. Each value is optimized only within $\{\text{1NN}(k=3),\text{kNN},\text{kNN-LOO}\}$, and Total comes from one estimator.}
  \label{fig:representation-ber-quality}
\end{figure*}

Figure~\ref{fig:representation-ber-quality} further tests whether BER-PEF is tied to a particular representation. Learned numeric representations reduce best Total on GeoLife/T-Drive from $0.237/0.075$ for RawTraj to $0.070/0.046$ for \textsc{SeqMLP} and $0.074/0.034$ for \textsc{UniMob}. Under time-plus-category context, \textsc{SeqMLP-C} and \textsc{UniMob-C} obtain $0.315/0.297$ and $0.297/0.263$, both below the $0.582/0.561$ of OneHot. For symbolic data, Frequency, Ngram-PCA, Linear-Ngram, \textsc{SeqMLP}, and \textsc{UniMob} all produce comparable LD/UD results under the same protocol. BER-PEF is therefore not tied to \textsc{UniMob}: the task-relevant structure retained by a representation affects BER quality, while multiple structured representations can enter the same evaluation protocol. This result also illustrates the role of predictability analysis in representation design. When a representation leads to poorer prediction performance and higher discrepancy, the limitation may arise from predictive information lost during representation rather than from the predictor alone. Predictability evaluation therefore provides a task-level criterion for comparing representations and diagnosing whether an input representation preserves information relevant to the prediction target.

\subsection{SOTA-Reference Robustness}

\begin{figure*}[!t]
  \centering
  \includegraphics[width=0.80\textwidth,trim=0 13.5bp 0 0,clip]{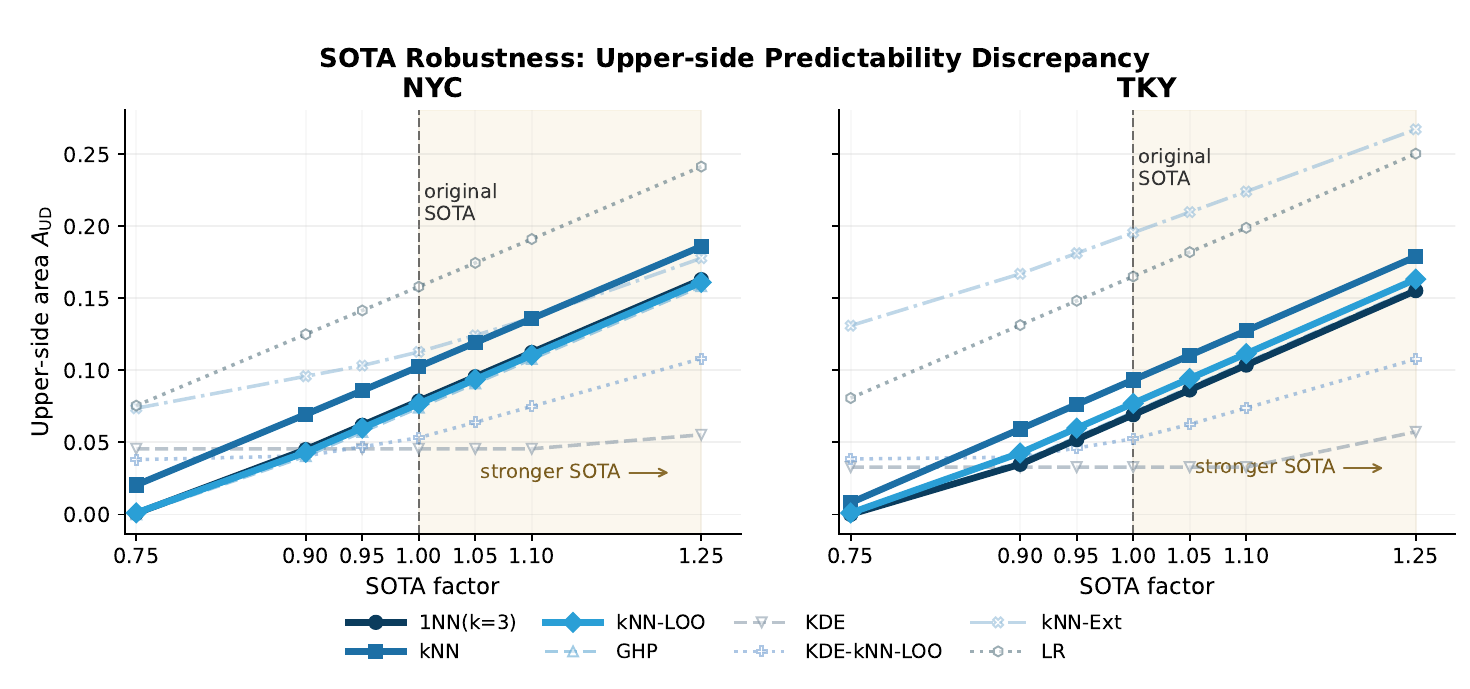}
  \setlength{\abovecaptionskip}{3pt}
  \caption{Upper-side discrepancy obtained by varying the SOTA-accuracy reference while keeping estimator outputs fixed. The horizontal axis is the SOTA factor, and lower values indicate closer agreement with the upper-side reference.}
  \label{fig:robustness}
\end{figure*}

The conservative predictability reference interval depends on the original-task SOTA accuracy $a_D^{\mathrm{SOTA}}$. To isolate reference uncertainty, we fix the estimator outputs from the symbolic-sequence experiment and vary only $a_D^{\mathrm{SOTA}}$ over factors $0.75$, $0.90$, $0.95$, $1.00$, $1.05$, $1.10$, and $1.25$. Each setting recomputes discrepancy without changing the perturbed samples, representation, labels, or estimator outputs. The main analysis focuses on the core estimators 1NN($k=3$), kNN, and kNN-LOO.

Figure~\ref{fig:robustness} shows that the absolute UD values change with the SOTA reference, while the core winner sets remain stable. On NYC, the core-only UD winner set contains kNN-LOO and 1NN($k=3$); on TKY, it contains 1NN($k=3$). The corresponding LD winners are kNN-LOO on NYC and 1NN($k=3$) on TKY. Reference uncertainty therefore changes the numerical discrepancy scale without altering the principal structure of the core estimator comparison.

\section{Conclusion}
\label{sec:conclusion}

We presented BER-PEF to address the absence of observable predictability ground truth and the resulting difficulty of comparing predictability estimators on real mobility data. BER-PEF defines the predictability limit as the complement of Bayes-optimal error and combines BER-based estimation with controlled perturbation evaluation through a shared reference interval and discrepancy measures. Its unified representation layer supports symbolic sequences, continuous trajectories, context-enhanced inputs, and multiple learned representations. Experiments show that the framework provides consistent evaluation across heterogeneous inputs. Overall, BER-PEF offers a unified, comparable, and verifiable path for predictability estimation and reliability analysis when observable predictability ground truth is unavailable.

\bibliographystyle{IEEEtran}
\bibliography{references}

\end{document}